\documentclass{article}

\usepackage[final]{nips_2018}

\usepackage[utf8]{inputenc} % allow utf-8 input
\usepackage[T1]{fontenc}    % use 8-bit T1 fonts
\usepackage{hyperref}       % hyperlinks
\usepackage{url}            % simple URL typesetting
\usepackage{booktabs}       % professional-quality tables
\usepackage{amsfonts}       % blackboard math symbols
\usepackage{nicefrac}       % compact symbols for 1/2, etc.
\usepackage{microtype}      % microtypography
\usepackage{amsmath}
\usepackage{graphicx}
\usepackage{listings}
\usepackage{subcaption}
\usepackage{color}
\usepackage{wrapfig}
\usepackage{tikz}
\usetikzlibrary{arrows.meta}

\title{Recurrent Relational Networks}

\author{Rasmus Berg Palm \\
Technical University of Denmark\\
Tradeshift \\
\texttt{rapal@dtu.dk}
\And Ulrich Paquet \\
DeepMind \\
\texttt{upaq@google.com}
\And Ole Winther \\
Technical University of Denmark \\
\texttt{olwi@dtu.dk}
}

\newcommand{\rbp}[1]{
}
\newcommand{\ulrich}[1]{
}

\begin{document}

\maketitle

\begin{abstract}
This paper is concerned with learning to solve tasks that require a chain of interdependent steps of relational inference,
like answering complex questions about the relationships between objects, or 
solving puzzles where the smaller elements of a solution mutually constrain each other.
We introduce the \emph{recurrent relational network},
a general purpose module that operates on a \emph{graph} representation of objects.
As a generalization of \cite{santoro2017simple}'s relational network,
it can augment any neural network model with the capacity to do many-step relational reasoning.
We achieve state of the art results on the bAbI textual question-answering dataset
with the recurrent relational network, consistently solving 20/20 tasks.
As bAbI is not particularly
challenging from a relational reasoning point of view,
we introduce Pretty-CLEVR, a new diagnostic dataset for relational reasoning.
In the Pretty-CLEVR set-up, we can vary the question to control for the number of relational reasoning steps that are required to obtain the answer.
Using Pretty-CLEVR, we probe the limitations of multi-layer perceptrons, relational and recurrent relational networks. 
Finally, we show how recurrent relational networks can learn to solve Sudoku puzzles from supervised training data, a challenging task requiring upwards of 64 steps of relational reasoning. We achieve state-of-the-art results amongst comparable methods by solving 96.6\% of the hardest Sudoku puzzles.
\end{abstract}

\section{Introduction}
%\rbp{Reorganized, so the flow is: 1) humans naturally do relational reasoning. Then example of relational reasoning so the reader know what we mean and can feel himself thinking relationally and thus buy the argument that it's natural 2) standard deep learning architectures can't do this. Since clearly they should, it's a major problem they don't. Thus, problem defined. 3) we show something universal, awesome, that can solve problem. Wuhu! Curiosity invoked.. but how do you do it? 4) scoping and ass covering: we only consider the reasoning aspect, yes the tasks are trivial for symbolic algorithms, but not for differentiable learned architectures and yes the term "relational reasoning" is loaded.}

A central component of human intelligence is the ability to abstractly reason about objects and their interactions \citep{spelke1995development,spelke2007core}. As an illustrative example,
consider solving a Sudoku. 
A Sudoku consists of 81 cells that are arranged in a 9-by-9 grid, which must be filled with digits 1 to 9 so that
each digit appears exactly once in each row,
column and 3-by-3 non-overlapping box, with a number of digits given \footnote{We invite the reader to solve the Sudoku in the supplementary material to appreciate the difficulty of solving a Sudoku in which 17 cells are initially filled.}.
To solve a Sudoku, one methodically reasons about the puzzle in terms of its cells and their interactions over many steps.
One tries placing digits in cells and see how that affects other cells, iteratively working toward a solution.
%In a representation, the cells (or objects) correspond to vertices in a graph, and each cell value is represented by a real-valued vector.
%By repeatedly considering neighboring vertices, these vectors are iteratively refined until the Sudoku is solved.

Contrast this with the canonical deep learning approach to solving problems, the multilayer perceptron (MLP), or multilayer convolutional neural net (CNN). These architectures take the entire Sudoku as an input and output the entire solution in a single forward pass, ignoring the inductive bias that objects exists in the world, and that they affect each other in a consistent manner. Not surprisingly these models fall short when faced with problems that require even basic relational reasoning \citep{lake2016building,santoro2017simple}. %\rbp{added to emphasize the contrast between relational/non relational and how stupid it is to solve relational problems with a non-relational model.}

The relational network of \citet{santoro2017simple} is an important first step towards a simple module for reasoning about objects and their interactions but it is limited to performing a single relational operation, and was evaluated on datasets that require a maximum of three steps of reasoning (which, surprisingly, can be solved by a single relational reasoning step as we show). 
Looking beyond relational networks, there is a rich literature on logic and reasoning in artificial intelligence and machine learning,
which we discuss in section \ref{related}.

Toward generally realizing the ability to methodically reason about objects and their interactions over many steps, this paper introduces a composite function, the \emph{recurrent relational network}. It serves as a modular component for many-step relational reasoning in end-to-end differentiable learning systems. It encodes the inductive biases that 1) objects exists in the world 2) they can be sufficiently described by properties 3) properties can change over time 4) objects can affect each other and 5) given the properties, the effects object have on each other is \emph{invariant to time}.
% These are powerful and intuitive inductive biases. As an example, we expect the suns gravitational pull on earth to be the same today, as it is in a year, given their mass and distance are the same.
%% Ulrich's comment. These two sentences don't aid the reader, and distracts from the main message. Especially, avoid the word "powerful", as it is overused (by Ilya S.) :)

%The modular component can be trained to answer questions that require multiple steps of relational reasoning about objects in simple scenes, and short stories.
%The same component can be trained to solve Sudoku puzzles. \rbp{rephrased, since we train on the states in pretty-clevr not the images.}
%There, it considers the repeated, increasingly restricted interaction between Sudoku cells, to iteratively narrow
%down representations of their contents, until a solution is reached. 

An important insight from the work of \citet{santoro2017simple} is to decompose a function
for relational reasoning into two components or ``modules'':
a perceptual front-end, which is tasked to recognize objects in the raw input and represent them as vectors,
and a relational reasoning module, which uses the representation to reason about the objects and their interactions.
Both modules are trained jointly end-to-end.
In computer science parlance, the relational reasoning module implements an \emph{interface}: it operates on a graph of nodes 
and directed edges, where the nodes are represented by real valued vectors, and is differentiable.
This paper chiefly develops the relational reasoning side of that interface.
% In this paper we'll only consider the relational reasoning side of that interface, since that is the part still in its infancy, and very good perceptual front-ends exists, e.g. convolutional networks for images, recurrent neural networks for text, etc. %\rbp{added scoping. To me that's the reason we define the interface in the first place. }

Some of the tasks we evaluate on can be efficiently and perfectly solved by hand-crafted algorithms that operate on the symbolic level. For example, 9-by-9 Sudokus can be solved in a fraction of a second with constraint propagation and search \citep{norvig2006solving} or with dancing links \citep{knuth2000dancing}. These symbolic algorithms are superior in every respect but one: they don't comply with the interface, as they are not differentiable and don't work with real-valued vector descriptions. They therefore cannot be used in a combined model with a deep learning perceptual front-end and learned end-to-end. %\rbp{rephrased from "by a recursive symbolic algorithms" since recursion is not required} but it clearly illustrate the basic components of the relational reasoning.
%\rbp{I fear that reviewers/readers will think: Bah, the tasks are trivially solved by normal algorithms, and then have that in mind for the next (complicated) section. While they're reading and trying to understand they'll be thinking, "Why make it so complicated when I can just solve this with e.g. constraint propagation?!" and be annoyed. I think we need to get ahead of that argument.}

Following \cite{santoro2017simple}, we use the term ``relational reasoning'' liberally for an object- and interaction-centric approach to problem solving. Although the term ``relational reasoning'' is similar to terms in other branches of science, like relational logic or first order logic, no direct parallel is intended.
%\ulrich{great! it covers our backs :)}

This paper considers many-step relational reasoning, a challenging task for deep learning architectures.
We develop a recurrent relational reasoning module, which constitutes our main contribution.
We show that it is a powerful architecture for many-step relational reasoning on three varied datasets, achieving state-of-the-art results on bAbI and Sudoku.
%\rbp{A first stab at a "contributions" paragraph. It feels a bit like repeating the abstract.}

\section{Recurrent Relational Networks}

We ground the discussion of a recurrent relational network in something familiar, solving a Sudoku puzzle.
A simple strategy works by
noting that if a certain Sudoku cell is given as a
``7'', one can safely remove ``7'' as an option from other cells in the same row, column and box.
In a message passing framework,
that cell needs to send a message to each other cell in the same row, column, and box,
broadcasting it's value as ``7'', and informing those cells not to take the value ``7''.
In an iteration $t$, these messages are sent simultaneously, in parallel, between all cells.
Each cell $i$ should then consider all incoming messages, and update its internal state $h_i^t$ to $h_i^{t+1}$.
With the updated state each cell should send out new messages, and the process repeats.

\paragraph{Message passing on a graph.}
The recurrent relational network will learn to pass messages on a \emph{graph}.
For Sudoku,
the graph has $i \in \{1,2,...,81\}$ nodes, one for each cell in the Sudoku.
Each node has an input feature vector $x_i$,
and edges to and from all nodes that are in the same row, column and box in the Sudoku.
The \emph{graph} is the input to the relational reasoning module,
and vectors $x_i$ would generally be the output of a perceptual front-end, for instance a convolutional neural network.
Keeping with our Sudoku example, each $x_i$ encodes the initial cell content (empty or given) and the row and column position of the cell.

At each step $t$ each node has a hidden state vector $h_i^t$,
which is initialized to the features, such that $h_i^0 = x_i$.
At each step $t$, each node sends a message to each of its neighboring nodes.
We define the message $m_{ij}^t$ from node $i$ to node $j$ at step $t$ by
\begin{align}
\label{eq:message}
    m_{ij}^t &= f\left(h_i^{t-1}, h_j^{t-1}\right) \ ,
\end{align}
where $f$, the message function, is a multi-layer perceptron.
This allows the network to learn what kind of messages to send.
In our experiments, MLPs with linear outputs were used.
Since a node needs to consider all the incoming messages we sum them with
\begin{align}
\label{eq:sum}
    m_j^t &= \sum_{i \in N(j)} m_{ij}^t \ ,
\end{align}
where $N(j)$ are all the nodes that have an edge into node $j$.
For Sudoku, $N(j)$ contains the nodes in the same row, column and box as $j$.
In our experiments, since the messages in \eqref{eq:message} are linear, this is similar to how log-probabilities are summed in belief propagation \citep{murphy1999loopy}.

\paragraph{Recurrent node updates.}
Finally we update the node hidden state via
\begin{align}
\label{eq:recurrence}
    h_j^t &= g\left(h_j^{t-1}, x_j, m_j^t\right) \ ,
\end{align}
where $g$, the node function, is another learned neural network.
The dependence on the previous node hidden state $h_j^{t-1}$ allows the network to iteratively work towards a solution instead of starting with a blank slate at every step.
Injecting the feature vector $x_j$ at each step like this allows the node function to focus on the messages from the other nodes instead of trying to remember the input.

%\begin{wrapfigure}[27]{r}{0.55\textwidth}
% \vspace{-10pt}
\begin{figure}[t]
\centering
\begin{tikzpicture}
  \begin{scope}[every node/.style={circle,draw}]
      \node[very thick, draw=green] (1) at (0,0) {$h_1^t$};
	  \node[draw=red] (x1) at (0,-1.25) {$x_1$};
      \node[draw=blue] (o1) at (-1.25,0) {$o_1^t$};
                       
      \node[very thick, draw=green] (2) at (4,0) {$h_2^t$};
      \node[draw=red] (x2) at (4,-1.25) {$x_2$};
      \node[draw=blue] (o2) at (5.25,0) {$o_2^t$};
      
      \node[very thick, draw=green] (3) at (2,3) {$h_3^t$};   
      \node[draw=red] (x3) at (1.116,3.884) {$x_3$};
      \node[draw=blue] (o3) at (2.884,3.884) {$o_3^t$};      
  \end{scope}
  \begin{scope}[>={Stealth[black]},
                every node/.style={fill=white, circle, pos=0.50, inner sep=1},
                every edge/.style={draw=black}]

      \path [->] (1) edge[dashed, out=140, in=110, looseness=8] (1);
      \path [->] (2) edge[dashed, out=60, in=30, looseness=8] (2);
      \path [->] (3) edge[dashed, loop above] (3);
      
      \path [->] (x1) edge (1);
      \path [->] (1) edge (o1);
      
      \path [->] (x2) edge (2);
      \path [->] (2) edge (o2);
      
      \path [->] (x3) edge (3);
      \path [->] (3) edge (o3);
      
      \path [->] (1) edge[bend left=20] node {$m_{12}^t$} (2);
      \path [->] (2) edge[bend left=20] node {$m_{21}^t$} (1);
      
      \path [->] (1) edge[bend left=20] node {$m_{13}^t$} (3);   
      \path [->] (3) edge[bend left=20] node {$m_{31}^t$} (1);
      
      \path [->] (2) edge[bend left=20] node {$m_{23}^t$} (3);  
      \path [->] (3) edge[bend left=20] node {$m_{32}^t$} (2);           

  \end{scope}
  \end{tikzpicture}
\caption{A \emph{recurrent relational network} on a fully connected graph with 3 nodes.
The nodes' hidden states $h_i^t$ are highlighted with green, the inputs $x_i$ with red, and the outputs $o_i^t$ with blue. The dashed lines indicate the recurrent connections.
Subscripts denote node indices and superscripts denote steps $t$.
For a figure of the same graph unrolled over 2 steps see the supplementary material. \label{fig:simple-graph}}
\end{figure}
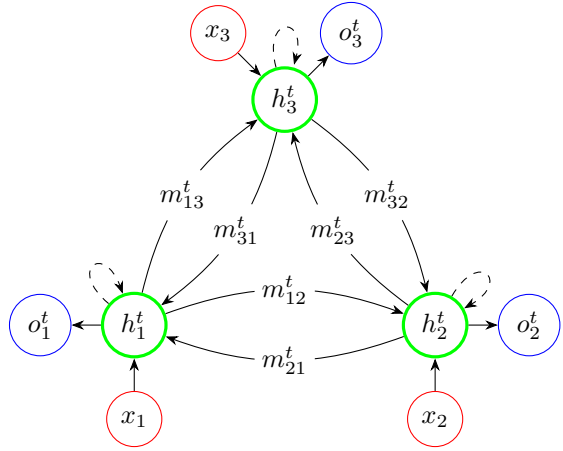
%\end{wrapfigure}

\paragraph{Supervised training.}
The above equations for sending messages and updating node states define a recurrent relational network's core. To train a recurrent relational network in a supervised manner to solve a Sudoku we introduce an output probability distribution over the digits 1-9 for each of the nodes in the graph.
The output distribution $o_i^t$ for node $i$ at step $t$ is given by
\begin{align}
    \label{eq:output}
    o_i^t &= r\left(h_i^t\right) \ ,
\end{align}
where $r$ is a MLP that maps the node hidden state to the output probabilities, e.g. using a softmax nonlinearity. Given the target digits $\mathbf{y} = \{y_1, y_2, ..., y_{81}\}$ the loss at step $t$, is then the sum of cross-entropy terms, one for each node: $l^t =  -\sum_{i=1}^{I} \log o_i^t\left[ y_i \right]$, where $o_i[y_i]$ is the $y_i$'th component of $o_i$.
Equations \eqref{eq:message} to \eqref{eq:output}
are illustrated in figure \ref{fig:simple-graph}.

%Given the target digit $y_i$ (1-9) for cell $i$, the cross-entropy node loss $l_i^t$ for node $i$ at step $t$ is
%
%\begin{align}
%    \label{eq:node_loss}
%    l_i^t &= -\log o_i^t\left[ y_i \right] \ ,
%\end{align}
%
%where the square brackets are used to indicate the $y_i$'th element of the vector.
%For a single Sudoku puzzle $\mathbf{x} = \{x_1, x_2, ..., x_{81}\}$ and its solution $\mathbf{y} = \{y_1, y_2, %..., y_{81}\}$ the total loss $\mathcal{L}\left(\bf{x}, \bf{y}\right)$ is the sum of losses computed recurrently %over all $I=81$ nodes and $T$ steps,
%\ulrich{The effect of the loss is getting lost in the equation below, as the dependence on trying to predict %the full Sudoku at each step is not clear. We should explain L in words below it.}
%
%\begin{align}
%    \label{eq:example_loss}
%    \mathcal{L}\left(\bf{x}, \bf{y}\right) &= \sum_{t=1}^T \sum_{i=1}^{I} l_i^t \ .
%\end{align}
%
%To train the network we minimize the total loss, with respect to the parameters of the functions $f$, $g$ and $r$ using stochastic gradient descent.

\paragraph{Convergent message passing.}
A distinctive feature of our proposed model is that we minimize the cross entropy between the output and target distributions at \emph{every step}.

At test time we only consider the output probabilities at the last step, but having a loss at every step during training is beneficial.
Since the target digits $y_i$ are constant over the steps, it encourages the network to learn a convergent message passing algorithm.
%As the message passing converges, the outputs will not change as $t$ goes to infinity. \rbp{Overly strong statement, will give reviewers ammo}
Secondly, it helps with the vanishing gradient problem.
%One potential issue with having a loss at every step is that it might force the network to learn a greedy algorithm that gets stuck in a local minima.
%However, the separate output function $r$ allows the node hidden states and messages to be different from the output probability distributions.
%The network therefore has the capacity to use a small part of the hidden state for retaining a current best guess, which can remain constant over several steps, and other parts of the hidden state for running a non-greedy multi-step algorithm.

%\ulrich{Rasmus, maybe we can work some of ``Sending messages for all nodes in parallel\ldots'' into the paragraph above?}

%\paragraph{Message passing schedule.}
%Sending messages for all nodes in parallel and summing all the incoming messages might seem like an unsophisticated approach that risk resulting in oscillatory behavior and drowning out the important messages.
%However, since the receiving node hidden state is an input to the message function, the receiving node can in a sense determine which messages it wishes to receive.
%As such, the sum can be seen as an implicit attention mechanism over the incoming messages.
%Similarly the network can learn an optimal message passing schedule, by ignoring messages based on the history and current state of the receiving and sending node.

\paragraph{Variations.}
If the edges are unknown, the graph can be assumed to be fully connected.
In this case the network will need to learn which objects interact with each other.
If the edges have attributes, $e_{ij}$, the message function in equation \ref{eq:message} can be modified such that $m_{ij}^t = f\left(h_i^{t-1}, h_j^{t-1}, e_{ij}\right)$.
If the output of interest is for the whole graph instead of for each node the output in equation \ref{eq:output} can be modified such that there's a single output $o^t = r\left(\sum_i h_i^t\right)$. The loss can be modified accordingly.

\section{Experiments}
Code to reproduce all experiments can be found at 
\href{https://github.com/rasmusbergpalm/recurrent-relational-networks}{github.com/rasmusbergpalm/recurrent-relational-networks}.
\subsection{bAbI question-answering tasks}
\begin{table}[!ht]
\caption{bAbI results. Trained jointly on all 20 tasks using the 10,000 training samples. Entries marked with an asterix are our own experiments, the rest are from the respective papers.}
\label{table:babi-results}
\centering
  \begin{tabular}{lccc}
  \toprule
  Method & N & Mean Error (\%) & Failed tasks (err. \textgreater 5\%) \\  
  \midrule
  \textit{RRN}* (this work) & $15$ & $\bf{0.46 \pm 0.77}$ & $\bf{0.13 \pm 0.35}$ \\
  SDNC \citep{rae2016scaling} & $15$ &  $6.4 \pm 2.5$ & $4.1 \pm 1.6$\\
  DAM \citep{rae2016scaling} & $15$ &  $8.7 \pm 6.4$ & $5.4 \pm 3.4$ \\
  SAM \citep{rae2016scaling} & $15$ &  $11.5 \pm 5.9$ & $7.1 \pm 3.4$\\  
  DNC \citep{rae2016scaling} & $15$ &  $12.8 \pm 4.7$ & $8.2 \pm 2.5$ \\
  NTM \citep{rae2016scaling} & $15$ &  $26.6 \pm 3.7$& $15.5 \pm 1.7$ \\
  LSTM \citep{rae2016scaling} & $15$ & $28.7 \pm 0.5$ & $17.1 \pm 0.8$ \\
  EntNet \citep{henaff2016tracking} & $5$ & $9.7 \pm 2.6$ & $5 \pm 1.2$ \\ 
  ReMo \citep{yang2018finding} & $1$ & $1.2$ & $1$ \\  
  RN \citep{santoro2017simple} & $1$ & N/A & $2$ \\
  MemN2N \citep{sukhbaatar2015end} & $1$ & $7.5$ & $6$ \\
  \bottomrule
  \end{tabular}  
\end{table}

bAbI is a text based QA dataset from Facebook \citep{weston2015towards} designed as a set of prerequisite tasks for reasoning. It consists of 20 types of tasks, with 10,000 questions each, including deduction, induction, spatial and temporal reasoning. Each question, e.g. ``Where is the milk?'' is preceded by a number of facts in the form of short sentences, e.g.~``Daniel journeyed to the garden. Daniel put down the milk.'' The target is a single word, in this case ``garden'', one-hot encoded over the full bAbI vocabulary of 177 words. A task is considered solved if a model achieves greater than 95\% accuracy. The most difficult tasks require reasoning about three facts.

To map the questions into a graph we treat the facts related to a question as the nodes in a fully connected graph up to a maximum of the last 20 facts. The fact and question sentences are both encoded by Long Short Term Memory (LSTM) \citep{hochreiter1997long} layers  with 32 hidden units each. We concatenate the last hidden state of each LSTM and pass that through a MLP. The output is considered the node features $x_i$. Following \citep{santoro2017simple} all edge features $e_{ij}$ are set to the question encoding. We train the network for three steps.
At each step, we sum the node hidden states and pass that through a MLP to get a single output for the whole graph. For details see the supplementary material.

Our trained network solves 20 of 20 tasks in 13 out of 15 runs. This is state-of-the-art and markedly more stable than competing methods. See table \ref{table:babi-results}. We perform ablation experiment to see which parts of the model are important, including varying the number of steps. We find that using dropout and appending the question encoding to the fact encodings is important for the performance. See the supplementary material for details.

Surprisingly, we find that we only need a single step of relational reasoning to solve all the bAbI tasks. This is surprising since the hardest tasks requires reasoning about three facts. It's possible that there are superficial correlations in the tasks that the model learns to exploit. Alternatively the model learns to compress all the relevant fact-relations into the 128 floats resulting from the sum over the node hidden states, and perform the remaining reasoning steps in the output MLP. Regardless, it appears multiple steps of relational reasoning are not important for the bAbI dataset.

\subsection{Pretty-CLEVR}

%\begin{figure}[ht]
%\centering
%\frame{\includegraphics[width=0.32\textwidth]{figures/1b.pdf}}
%\frame{\includegraphics[width=0.32\textwidth]{figures/2.pdf}}
%\frame{\includegraphics[width=0.32\textwidth]{figures/3.pdf}}
%\caption{Three samples of the Pretty-CLEVR diagnostic dataset. Each sample has 128 questions associated, exhibiting varying levels of relational reasoning difficulty. For the leftmost sample the solution to the question: "green, 3 jumps", which is "plus", is shown with arrows.}
%\label{fig:pretty-dataset}
%\end{figure}

Given that bAbI did not require multiple steps of relational reasoning and in order to test our hypothesis that our proposed model is better suited for tasks requiring more steps of relational reasoning we create a diagnostic dataset ``Pretty-CLEVER''. It can be seen as an extension of the ``Sort-of-CLEVR'' data set by \citep{santoro2017simple} which has questions of a non-relational and relational nature. ``Pretty-CLEVR'' takes this a step further and has non-relational questions as well as questions requiring \textit{varying} degrees of relational reasoning.

\begin{figure}[ht]
\centering
\captionsetup{position=bottom}
\subcaptionbox{Samples.\label{fig:pretty:samples}}[.24\textwidth]
{
\frame{\includegraphics[width=.24\textwidth]{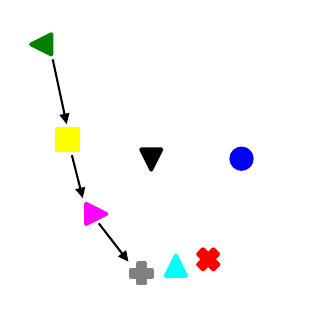}}
\frame{\includegraphics[width=.24\textwidth]{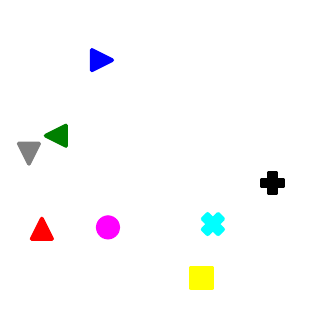}}
}
\subcaptionbox{Results.\label{fig:pretty:results}}
{\includegraphics[width=0.72\textwidth]{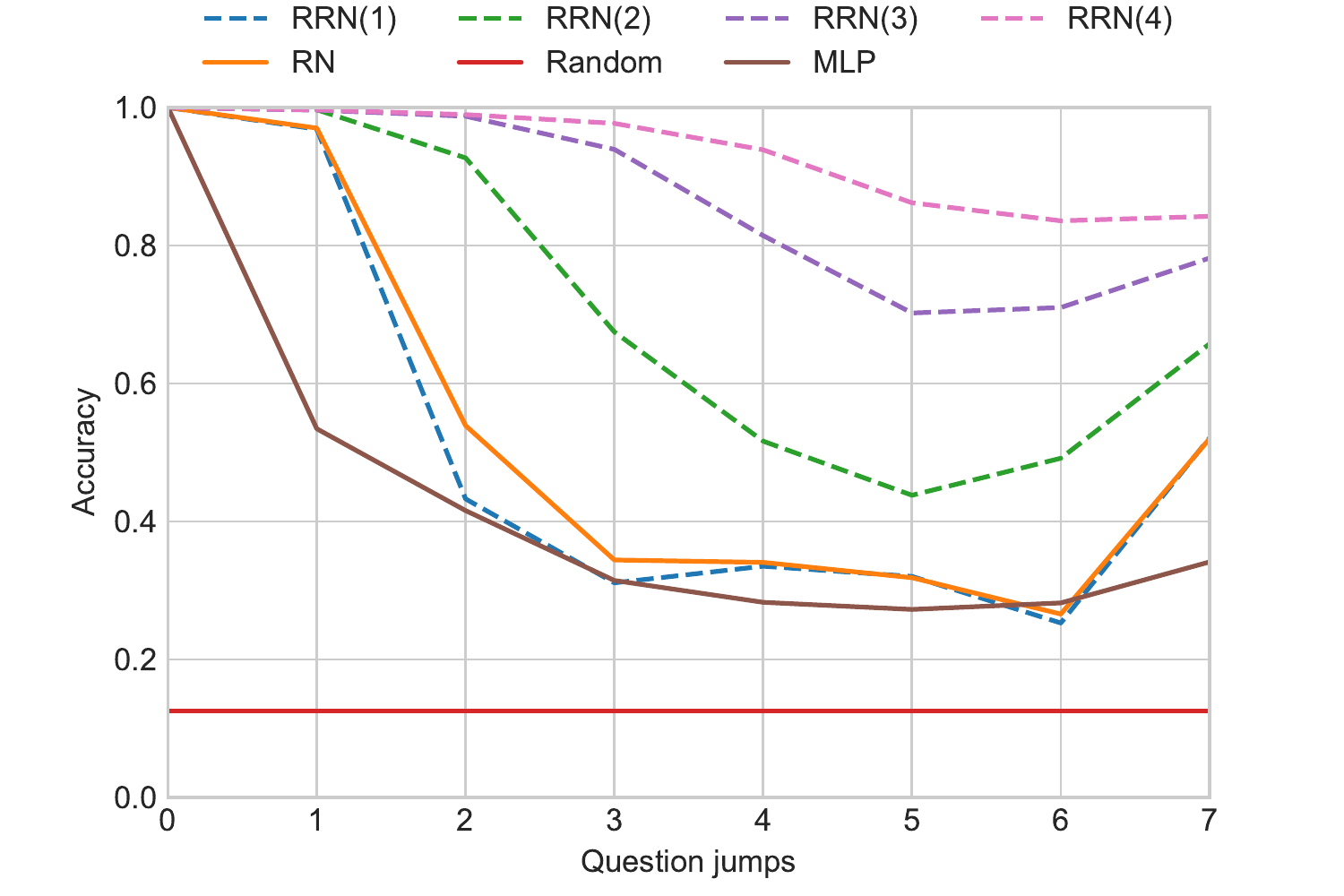}}
\caption{\ref{fig:pretty:samples} Two samples of the Pretty-CLEVR diagnostic dataset. Each sample has 128 questions associated, exhibiting varying levels of relational reasoning difficulty. For the topmost sample the solution to the question:
``green, 3 jumps'', which is ``plus'', is shown with arrows. \ref{fig:pretty:results} Random corresponds to picking one of the eight possible outputs at random (colors or shapes, depending on the input). The RRN is trained for four steps but since it predicts at each step we can evaluate the performance for each step. The the number of steps is stated in parentheses. \rbp{compacted the two pretty figures into one.}}
\end{figure}

Pretty-CLEVR consists of scenes with eight colored shapes and associated questions. Questions are of the form: ``Starting at object X which object is N jumps away?''. Objects are uniquely defined by their color or shape. If the start object is defined by color, the answer is a shape, and vice versa. Jumps are defined as moving to the closest object, without going to an object already visited. See figure \ref{fig:pretty:samples}. 
Questions with zero jumps are non-relational and correspond to: ``What color is shape X?'' or ``What shape is color X?''. We create 100,000 random scenes, and 128 questions for each (8 start objects, 0-7 jumps, output is color or shape), resulting in 12.8M questions. We also render the scenes as images. The ``jump to nearest'' type question is chosen in an effort to eliminate simple correlations between the scene state and the answer. It is highly non-linear in the sense that slight differences in the distance between objects can cause the answer to change drastically. It is also asymmetrical, i.e. if the question ``x, n jumps'' equals ``y'', there is no guarantee that ``y, n jumps'' equals ``x''. We find it is a surprisingly difficult task to solve, even with a powerful model such as the RRN. We hope others will use it to evaluate their relational models.\footnote{Pretty-CLEVR is available online as part of the code for reproducing experiments.}

Since we are solely interested in examining the effect of multiple steps of relational reasoning we train on the state descriptions of the scene. We consider each scene as a fully connected undirected graph with 8 nodes. The feature vector for each object consists of the position, shape and color. We encode the question as the start object shape or color and the number of jumps. As we did for bAbI we concatenate the question and object features and pass it through a MLP to get the node features $x_i$. To make the task easier we set the edge features to the euclidean distance between the objects. We train our network for four steps and compare to a single step relational network and a baseline MLP that considers the entire scene state, all pairwise distances, and the question as a single vector. For details see the supplementary material.

Mirroring the results from the ``Sort-of-CLEVR'' dataset the MLP perfectly solves the non-relational questions, but struggle with even single jump questions and seem to lower bound the performance of the relational networks. The relational network solves the non-relational questions as well as the ones requiring a single jump, but the accuracy sharply drops off with more jumps. This matches the performance of the recurrent relational network which generally performs well as long as the number of steps is greater than or equal to the number of jumps. See fig \ref{fig:pretty:results}. It seems that, despite our best efforts, there are spurious correlations in the data such that questions with six to seven jumps are easier to solve than those with four to five jumps.

\subsection{Sudoku} \label{sec:sudoku}
We create training, validation and testing sets totaling 216,000 Sudoku puzzles with a uniform distribution of givens between 17 and 34. We consider each of the 81 cells in the 9x9 Sudoku grid a node in a graph, with edges to and from each other cell in the same row, column and box. The node features $x_i$ are the output of a MLP which takes as input the digit for the cell (0-9, 0 if not given), and the row and column position (1-9). Edge features are not used. We run the network for 32 steps and at every step the output function $r$ maps each node hidden state to nine output logits corresponding to the nine possible digits. For details see the supplementary material.

\begin{figure}[!ht]
\centering
{\includegraphics[width=1.0\textwidth]{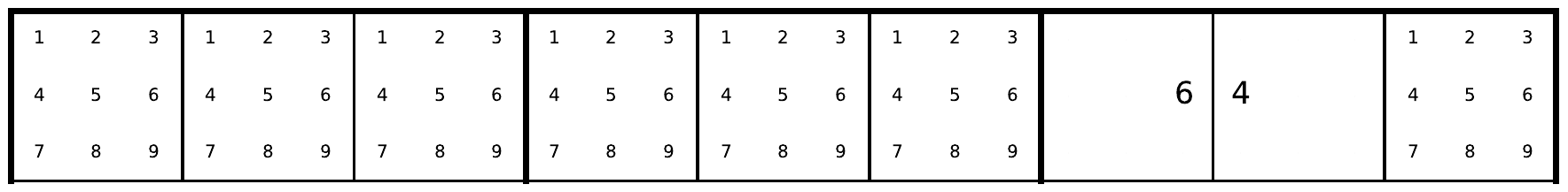}}
{\includegraphics[width=1.0\textwidth]{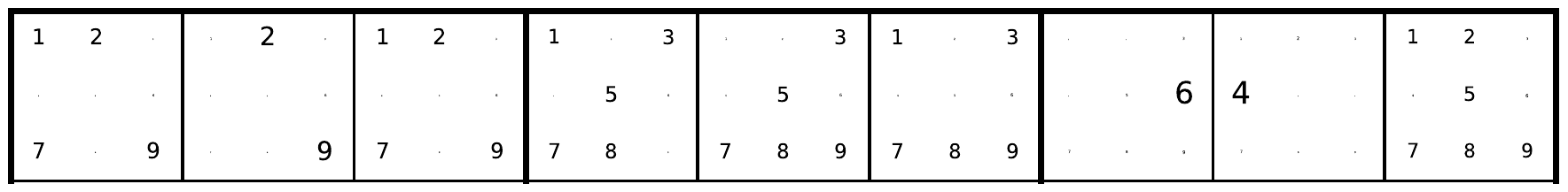}}
{\includegraphics[width=1.0\textwidth]{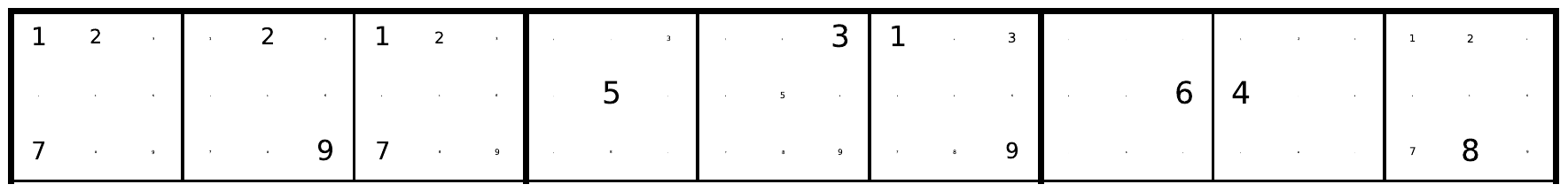}}
{\includegraphics[width=1.0\textwidth]{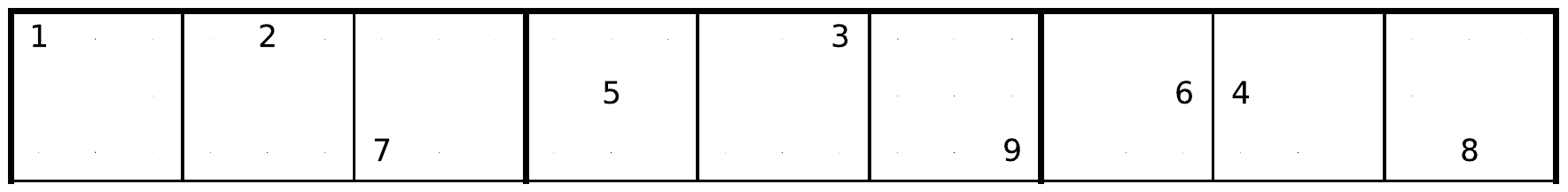}}
\caption{Example of how the trained network solves part of a Sudoku. Only the top row of a full 9x9 Sudoku is shown for clarity. From top to bottom steps 0, 1, 8 and 24 are shown. See the supplementary material for a full Sudoku. Each cell displays the digits 1-9 with the font size scaled (non-linearly for legibility) to the probability the network assigns to each digit. Notice how the network eliminates the given digits 6 and 4 from the other cells in the first step. Animations showing how the trained network solves Sodukos, including a failure case can be found at \href{https://imgur.com/a/ALsfB}{imgur.com/a/ALsfB}.}\label{fig:steps}
\end{figure}

Our network learns to solve 94.1\% of even the hardest 17-givens Sudokus after 32 steps. We only consider a puzzled solved if all the digits are correct, i.e.~no partial credit is given for getting individual digits correct. For more givens the accuracy (fraction of test puzzles solved) quickly approaches 100\%. Since the network outputs a probability distribution for each step, we can visualize how the network arrives at the solution step by step. For an example of this see figure \ref{fig:steps}.

\begin{figure}[ht]
\vspace{-10pt}
\centering
\includegraphics[width=1.0\textwidth]{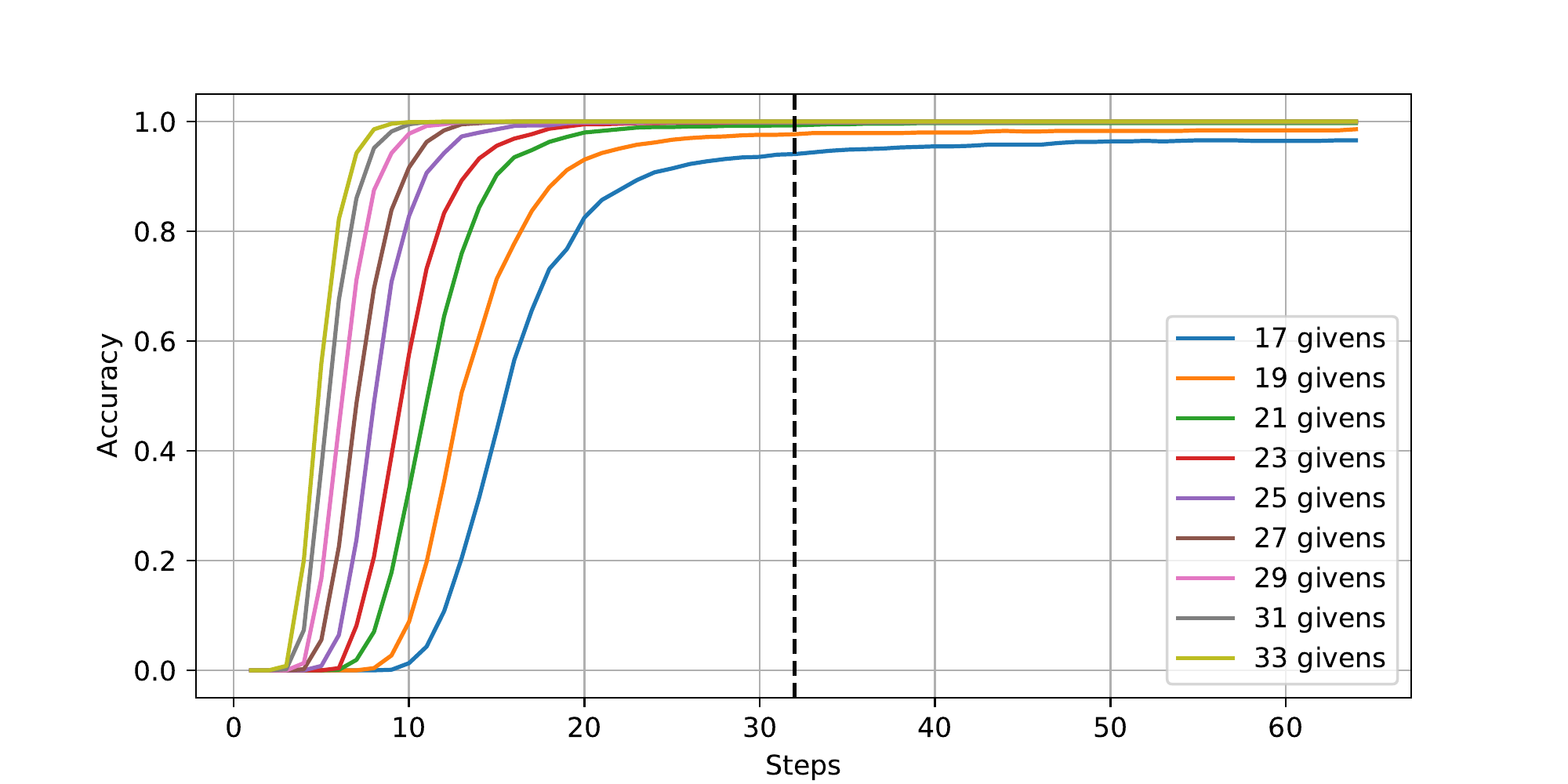}
\caption{Fraction of test puzzles solved as a function of number of steps. Even simple Sudokus with 33 givens require about 10 steps of relational reasoning to be solved. The dashed vertical line indicates the 32 steps the network was trained for. The network appears to have learned a convergent relational reasoning algorithm such that more steps beyond 32 improve on the hardest Sudokus.}
\label{fig:sudoku-acc}
\end{figure}

To examine our hypothesis that multiple steps are required we plot the accuracy as a function of the number of steps. See figure \ref{fig:sudoku-acc}. We can see that even simple Sudokus with 33 givens require upwards of 10 steps of relational reasoning, whereas the harder 17 givens continue to improve even after 32 steps. Figure \ref{fig:sudoku-acc} also shows that the model has learned a convergent algorithm. The model was trained for 32 steps, but seeing that the accuracy increased with more steps, we ran the model for 64 steps during testing. At 64 steps the accuracy for the 17 givens puzzles increases to 96.6\%.

We also examined the importance of the row and column features by multiplying the row and column embeddings by zero and re-tested our trained network. At 64 steps with 17 givens, the accuracy changed to 96.7\%. It thus seems the network does not use the row and column position information to solve the task.

We compare our network to several other differentiable methods.
See table \ref{table:sudoku-results}.
We train two relational networks: a node and a graph centric. For details see the supplementary material. Of the two, the node centric was considerably better. The node centric correspond exactly to our proposed network with a single step, yet fails to solve any Sudoku. This shows that multiple steps are crucial for complex relational reasoning.
Our network outperforms loopy belief propagation, with parallel and random messages passing updates \citep{bauke2008passing}.
It also outperforms a version of loopy belief propagation modified specifically for solving Sudokus that uses 250 steps, Sinkhorn balancing every two steps and iteratively picks the most probable digit \citep{khan2014solving}.
We also compare to learning the messages in parallel loopy BP as presented in \citet{lin2015deeply}.
We tried a few variants including a single step as presented and 32 steps with and without a loss on every step, but could not get it to solve any 17 given Sudokus.
Finally we outperform \citet{park2016can} which treats the Sudoku as a 9x9 image, uses 10 convolutional layers, iteratively picks the most probable digit, and evaluate on easier Sudokus with 24-36 givens.
We also tried to train a version of our network that only had a loss at the last step.
It was harder to train, performed worse and didn't learn a convergent algorithm.

\begin{table}[!ht]
\caption{Comparison of methods for solving Sudoku puzzles. Only methods that are differentiable are included in the comparison. Entries marked with an asterix are our own experiments, the rest are from the respective papers. }
\label{table:sudoku-results}
    \centering
	\begin{tabular}{lccc}
    \toprule
        Method & Givens & Accuracy \\ 
        \midrule        
		\emph{Recurrent Relational Network}* (this work)             & 17    & \textbf{96.6}\%  \\
        Loopy BP, modified  \citep{khan2014solving}                 & 17    & 92.5\% \\
        Loopy BP, random \citep{bauke2008passing}                   & 17    & 61.7\% \\
        Loopy BP, parallel \citep{bauke2008passing}                 & 17    & 53.2\% \\
        Deeply Learned Messages* \citep{lin2015deeply}               & 17    & 0\% \\
        Relational Network, node* \citep{santoro2017simple}          & 17    & 0\% \\
        Relational Network, graph* \citep{santoro2017simple}         & 17    & 0\% \\
        Deep Convolutional Network \citep{park2016can}              & 24-36 & 70\% \\
        \bottomrule
    \end{tabular}    
\end{table}

\subsection{Age arithmetic}
Anonymous reviewer 2 suggested the following task which we include here.
The task is to infer the age of a person given a single absolute age and a set of age differences, e.g. ``Alice is 20 years old. Alice is 4 years older than Bob. Charlie is 6 years younger than Bob. How old is Charlie?''. Please see the supplementary material for details on the task and results.

\section{Discussion} 

\rbp{This section could use some Ulrich magic :)}

We have proposed a general relational reasoning model for solving tasks requiring an order of magnitude more complex relational reasoning than the current state-of-the art. BaBi and Sort-of-CLEVR require a few steps, Pretty-CLEVR requires up to eight steps and Sudoku requires more than ten steps.
Our relational reasoning module can be added to any deep learning model to add a powerful relational reasoning capacity.
We get state-of-the-art results on Sudokus solving 96.6\% of the hardest Sudokus with 17 givens.
We also markedly improve state-of-the-art on the BaBi dataset solving 20/20 tasks in 13 out of 15 runs with a single model trained jointly on all tasks.

One potential issue with having a loss at every step is that it might encourage the network to learn a greedy algorithm that gets stuck in a local minima.
However, the output function $r$ separates the node hidden states and messages from the output probability distributions.
The network therefore has the capacity to use a small part of the hidden state for retaining a current best guess, which can remain constant over several steps, and other parts of the hidden state for running a non-greedy multi-step algorithm.

Sending messages for all nodes in parallel and summing all the incoming messages might seem like an unsophisticated approach that risk resulting in oscillatory behavior and drowning out the important messages.
However, since the receiving node hidden state is an input to the message function, the receiving node can in a sense determine which messages it wishes to receive.
As such, the sum can be seen as an implicit attention mechanism over the incoming messages.
Similarly the network can learn an optimal message passing schedule, by ignoring messages based on the history and current state of the receiving and sending node.

\section{Related work} \label{related}
Relational networks \citep{santoro2017simple} and interaction networks \citep{battaglia2016interaction} are the most directly comparable to ours.
These models correspond to using a single step of equation \ref{eq:recurrence}.
Since it only does one step it cannot naturally do complex multi-step relational reasoning.
In order to solve the tasks that require more than a single step it must compress all the relevant relations into a fixed size vector, then perform the remaining relational reasoning in the last forward layers.
Relational networks, interaction networks and our proposed model can all be seen as an instance of Graph Neural Networks \citep{scarselli2009graph,gilmer2017neural}.

Graph neural networks with message passing computations go back to \cite{scarselli2009graph}. However, there are key differences that we found important for implementing stable multi-step relational reasoning. Including the node features $x_j$ at every step in eq. \ref{eq:recurrence} is important to the stability of the network. \cite{scarselli2009graph}, eq. 3 has the node features, $l_n$, inside the message function. \citet{battaglia2016interaction} use an $x_j$ in the node update function, but this is an external driving force. \citet{sukhbaatar2016learning} also proposed to include the node features at every step.
Optimizing the loss at every step in order to learn a convergent message passing algorithm is novel to the best of our knowledge. \cite{scarselli2009graph} introduces an explicit loss term to ensure convergence. \citet{ross2011learning} trains the inference machine predictors on every step, but there are no hidden states; the node states are the output marginals directly, similar to how belief propagation works.

Our model can also be seen as a completely learned message passing algorithm.
Belief propagation is a hand-crafted message passing algorithm for performing exact inference in directed acyclic graphical models.
If the graph has cycles, one can use a variant, loopy belief propagation, but it is not guaranteed to be exact, unbiased or converge.
Empirically it works well though and it is widely used \citep{murphy1999loopy}.
Several works have proposed replacing parts of belief propagation with learned modules \citep{heess2013learning, lin2015deeply}.
Our work differs by not being rooted in loopy BP, and instead learning all parts of a general message passing algorithm.
\citet{ross2011learning} proposes Inference Machines which ditch the belief propagation algorithm altogether and instead train a series of regressors to output the correct marginals by passing messages on a graph.
\citet{wei2016convolutional} applies this idea to pose estimation using a series of convolutional layers and \citet{deng2016structure} introduces a recurrent node update for the same domain.

There is rich literature on combining symbolic reasoning and logic with sub-symbolic distributed representations which goes all the way back to the birth of the idea of parallel distributed processing \citet{mcculloch1943logical}.
See \citep{raedt2016statistical, besold2017neural} for two recent surveys.
Here we describe only a few recent methods.
\citet{serafini2016learning} introduces the Logic Tensor Network (LTN) which describes a first order logic in which symbols are grounded as vector embeddings, and predicates and functions are grounded as tensor networks.
The embeddings and tensor networks are then optimized jointly to maximize a fuzzy satisfiability measure over a set of known facts and fuzzy constraints.
%In \citet{donadello2017logic} the LTN is used to improve on a Semantic Image Interpretation task by incorporating fuzzy prior constraints, e.g. cats usually have tails.
\citet{vsourek2015lifted} introduces the Lifted Relational Network which combines relational logic with neural networks by creating neural networks from lifted rules and training examples, such that the connections between neurons created from the same lifted rules shares weights.
Our approach differs fundamentally in that we do not aim to bridge symbolic and sub-symbolic methods.
Instead we stay completely in the sub-symbolic realm.
We do not introduce or consider any explicit logic, aim to discover (fuzzy) logic rules, or attempt to include prior knowledge in the form of logical constraints.

\citet{amos2017optnet} Introduces OptNet, a neural network layer that solve quadratic programs using an efficient differentiable solver.
OptNet is trained to solve 4x4 Sudokus amongst other problems and beats the deep convolutional network baseline as described in \citet{park2016can}.
Unfortunately we cannot compare to OptNet directly as it has computational issues scaling to 9x9 Sudokus (Brandon Amos, 2018, personal communication).

\citet{sukhbaatar2016learning} proposes the Communication Network (CommNet) for learning multi-agent cooperation and communication using back-propagation. It is similar to our recurrent relational network, but differs in key aspects. The messages passed between all nodes at a given step are the same, corresponding to the average of all the node hidden states. Also, it is not trained to minimize the loss on every step of the algorithm.

\subsubsection*{Acknowledgments}
We'd like to thank the anonymous reviewers for the valuable comments and suggestions, especially reviewer 2 who suggested the age arithmetic task.
This research was supported by the NVIDIA Corporation with the donation of TITAN X GPUs.

\bibliography{refs}
\bibliographystyle{plainnat}

\section{Supplementary Material}

\subsection{bAbI experimental details}
Unless otherwise specified we use 128 hidden units for all layers and all MLPs are 3 ReLU layers followed by a linear layer.

We compute each node feature vector as 
\begin{align*}
x_i = \text{MLP}(\text{concat}(\text{last}(\text{LSTM}_S(s_i)), \text{last}(\text{LSTM}_Q(q)), \text{onehot}(p_i + o)))
\end{align*}
where $s_i$ is fact $i$, $q$ is the question, $p_i$ is the sentence position (1-20) of fact $i$ and $o$ is a random offset per question (1-20), such that the onehot output is 40 dimensional. The offset is constant for all facts related to a single question to avoid changing the relative order of the facts. The random offset prevents the network from memorizing the position of the facts and rather reason about their ordering.
Our message function $f$ is a MLP. Our node function $g$ uses an LSTM over reasoning steps
\begin{align*}
	h_j^t, s_j^t = \text{LSTM}_G(\text{MLP}(\text{concat}(x_j, m_{j}^t)), s_j^{t-1}) \ ,
\end{align*}
where $s_j^t$ is the cell state of the LSTM for unit $j$ at time $t$. $s_j^0$ is initialized to zero.

We run our network for three steps. To get a graph level output, we use a MLP over the sum of the node hidden states, $o^t = \text{MLP}\left(\sum_i h_i^t\right)$ with 3 layers, the final being a linear layer that maps to the output logits. The last two layers uses dropout of 50\%. We train and validate on all 20 tasks jointly using the 9,000 training and 1,000 validation samples defined in the \texttt{en\_valid\_10k} split. We use the Adam optimizer with a batch size of 512, a learning rate of 2e-4 and L2 regularization with a rate of 1e-5. We train for 5M gradient steps.

\subsection{bAbI ablation experiments}

To test which parts of the proposed model is important to solving the bAbI tasks we perform ablation experiments. One of the main differences between the relational network and our proposed model, aside from the recurrent steps, is that we encode the sentences and question together. We ablate the model in two ways to test how important this is. 1) Using a single linear layer instead of the 4-layer MLP baseline, and 2) Not encoding them together. In this case the node hidden states are initialized to the fact encodings. We found dropout to be important, so we also perform an ablation experiment without dropout. We run each ablation experiment eight times. We also do pseudo-ablation experiments with fewer steps by measuring at each step of the RRN. See table \ref{table:babi-ablate}.

\begin{table}[!ht]
\begin{centering}
  \begin{tabular}{lcccc}
  Model & Runs & Mean Error (\%) & Failed tasks (err. \textgreater 5\%) & Mean error @ 1M updates (\%)\\  
  \midrule
  Baseline, 3 steps & 15 & $0.46 \pm 0.77$ & $0.13 \pm 0.35$ & $1.83 \pm 1.06$ \\
  Baseline, 2 steps & 15 & $0.46 \pm 0.76$ & $0.13 \pm 0.35$ & $1.83 \pm 1.06$ \\
  Baseline, 1 step & 15 & $0.48 \pm 0.79$ & $0.13 \pm 0.35$ & $1.84 \pm 1.06$ \\
  linear encoding & 8 & $\bf{0.20 \pm 0.01}$ & $\bf{0 \pm 0}$ & $\bf{0.63 \pm 0.69}$ \\
  no encoding & 8 & $0.53 \pm 0.91$ & $0.13 \pm 0.35$ & $2.39 \pm 1.73$ \\
  no dropout & 8 & $1.74 \pm 1.28$ & $0.63 \pm 0.52$ & $2.57 \pm 0.95$
  \end{tabular}
  \caption{BaBi ablation results. \label{table:babi-ablate}}
  \end{centering}
\end{table}

\subsection{Pretty-CLEVR experimental details}
Our setup for Pretty-CLEVR is a bit simpler than for bAbI. Unless otherwise specified we use 128 hidden units for all hidden layers and all MLPs are 1 ReLU layer followed by a linear layer.

We compute each node feature vector $x_i$ as
\begin{align*}
o_i &= \text{concat}(p_i, \text{onehot}(c_i), \text{onehot}(m_i)) \\
q &= \text{concat}(\text{onehot}(s), \text{onehot}(n)) \\
x_i &= \text{MLP}(\text{concat}(o_i, q))
\end{align*}
where 
$p_i \in [0,1]^2$ is the position, 
$\mathbb{N}^n \equiv \{0,...,n-1\}$, 
$c_i \in \mathbb{N}^8$ is the color, 
$m_i \in \mathbb{N}^8$ is the marker, 
$s \in \mathbb{N}^{16}$ is the marker or color of the start object, and 
$n \in \mathbb{N}^8$ is the number of jumps.

Our message function $f$ is a MLP. Our node function $g$ is,
\begin{align*}
	h_j^t = \text{MLP}(\text{concat}(h_j^{t-1}, x_j, m_{j}^t))
\end{align*}

Our output function $r$ is a MLP with a dropout fraction of 0.5 in the penultimate layer. The last layer has 16 hidden linear units. We run our recurrent relational network for 4 steps.

We train on the 12.8M training questions, and augment the data by scaling and rotating the scenes randomly. We use separate validation and test sets of 128.000 questions each. We use the Adam optimizer with a learning rate of 1e-4 and train for 10M gradient updates with a batch size of 128.

The baseline RN is identical to the described RRN, except it only does a single step of relational reasoning.

The baseline MLP takes the entire scene state, $\mathbf{x}$, as an input, such that
\begin{align*}
	\mathbf{x} = \text{concat}(o_0, ..., o_7, d_{00}, ..., d_{77}, q)
\end{align*}
where $d_{ij} \in \mathbb{R}$ is the euclidean distance from object $i$ to $j$.

The baseline MLP has 4 ReLu layers with 256 hidden units, with dropout of 0.5 on the last layer, followed by a linear layer with 16 hidden units. The baseline MLP has 87\% more parameters than the RRN and RN (261,136 vs 139,536).

\subsection{Sudoku dataset}
To generate our dataset the starting point is the collection of 49,151 unique 17-givens puzzles gathered by \citet{royle2016minimum} which we solve using the solver from \citet{norvig2006solving}.
Then we split the puzzles into a test, validation and training \emph{pool}, with 10,000, 1,000 and 38,151 samples respectively.
To generate the \emph{sets} we train, validate and test on we do the following: for each $n \in \{0,...,17\}$ we sample $k$ puzzles from the respective pool, with replacement. For each sampled puzzle we add $n$ random digits from the solution. We then swap the digits according to a random permutation, e.g. $1 \to 5$, $2 \to 3$, etc. The resulting puzzle is added to the respective set. For the test, validation and training sets we sample $k=1,000$, $k=1,000$ and $k=10,000$ puzzles in this way.

\subsection{Sudoku experimental details}
Unless otherwise specified we use 96 hidden units for all hidden layers and all MLPs are 3 ReLU layers followed by a linear layer.

Denote the digit for cell $j$ $d_j$ (0-9, 0 if not given), and the row and column position $\text{row}_j$ (1-9) and $\text{column}_j$ (1-9) respectively..
The node features are then
\begin{align*}
	x_j = \text{MLP}(\text{concat}(\text{embed}(d_j),\text{embed}(\text{row}_j),\text{embed}(\text{column}_j)))
\end{align*}
where each $\text{embed}$ is a 16 dimensional learned embedding. We could probably have used one-hot encoding instead of the embeddings, embedding was just the first thing we tried. Edge features were not used.
The message function $f$ is an MLP. The node function $g$, is identical to the setup for bAbI, i.e.
\begin{align*}
	h_j^t, s_j^t = \text{LSTM}_G(\text{MLP}(\text{concat}(x_j, m_{j}^t)), s_j^{t-1}) \ .
\end{align*}
The LSTM cell state is initialized to zeros.

The output function $r$ is a linear layer with nine outputs to produce the output logits $o_i^t$. 
We run the network for 32 steps with a loss on every step. We train the model for 300.000 gradient updates with a batch size of 256 using Adam with a learning rate of 2e-4 and L2 regularization of 1e-4 on all weight matrices.

\subsection{Sudoku relational network baseline details}
The node centric corresponds exactly to a single step of our network.
The graph centric approach is closer to the original relational network.
It does one step of relational reasoning as our network, then sums all the node hidden states. The sum is then passed through a 4 layer MLP with $81\cdot9$ outputs, one for each cell and digit. The graph centric model has larger hidden states of 256 in all layers to compensate somewhat for the sum squashing the entire graph into a fixed size vector. Otherwise both networks are identical to our network. The graph centric has over 4 times as many parameters as our model (944,874 vs. 201,194) but performs worse than the node centric.

\subsection{Age arithmetic task details}
We generated all 262,144 unique trees with 8 nodes and split them 90\%/10\% into training and test graphs. The nodes represent the persons, and the edges which age differences will be given to the network. During training and testing we sample a batch of graphs from the respective set and sample 8 random ages (0-99) for each. We compute the absolute difference as well as the sign for each edge in the graphs. This gives us 7 relative facts on the form “person A (0-7), person B (0-7), younger/older (-1,1), absolute age difference (0-99)”. Then we add the final fact which is the age of one of the nodes at random, e.g. “3, 3, 0, 47”, using the zero sign to indicate this fact is absolute and not relative. The question is the age of one of the persons at random (0-7). For each graph we compute the shortest path from the anchor person to the person in question. This is the minimum number of arithmetic computations that must be performed to infer the persons age from the given facts.

\begin{figure}[!ht]
\centering
\includegraphics[width=1.0\textwidth]{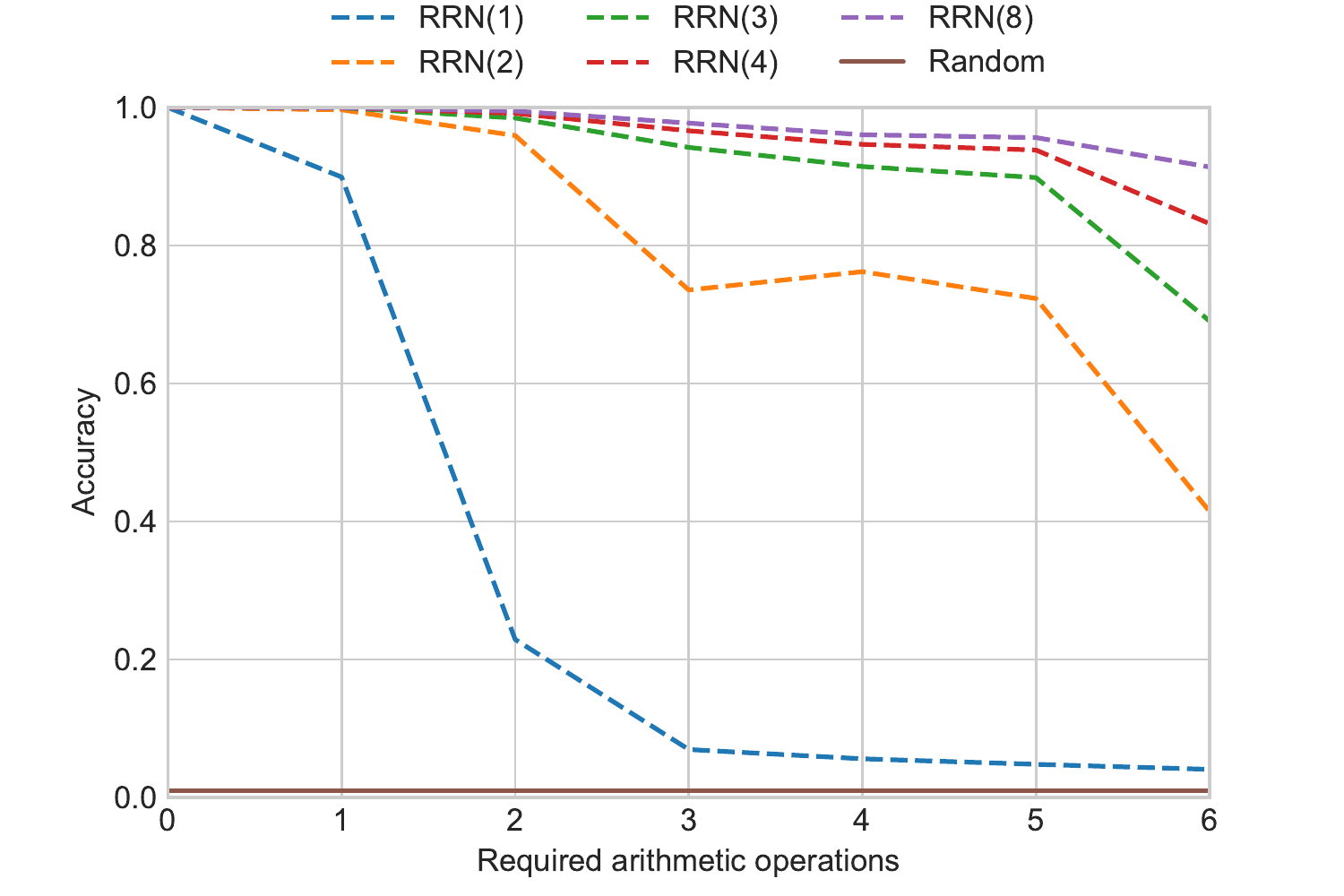}
\caption{Results for the age arithmetic task. The number in parenthesis indicate how many steps the RRN was run during testing. Random corresponds to picking one of the 100 possible ages randomly.
\label{fig:age-results}}
\end{figure}

The 8 facts (1 anchor, 7 relative) are given to the network as a fully connected graph of 8 nodes. Note, this graph is different from the tree used to generate the facts. The network never sees the tree. The input vector for each fact are the four fact integers and the question integer one-hot encoded and concatenated. We use the same architecture as for the bAbI experiments except all MLPs are 3 dense layers with 128 ReLu units followed by one linear layer. We train the network for 8 steps, and test it for each step. See figure \ref{fig:age-results} for results. 

\subsection{Unrolled recurrent relational network}

\begin{figure}[!ht]
\centering
\begin{tikzpicture} [scale = 1.0]
  \begin{scope}[every node/.style={circle,draw}]
      \node[draw=red] (x01) at (0,-1.25) {$x_1$};
      \node[draw=red] (x02) at (0,2.75) {$x_2$};
      \node[draw=red] (x03) at (0,6.75) {$x_3$};

      \node[draw=red] (x11) at (5,-1.25) {$x_1$};
      \node[draw=red] (x12) at (5,2.75) {$x_2$};
      \node[draw=red] (x13) at (5,6.75) {$x_3$};

      \node[draw=red] (x21) at (10,-1.25) {$x_1$};
      \node[draw=red] (x22) at (10,2.75) {$x_2$};
      \node[draw=red] (x23) at (10,6.75) {$x_3$};

      % step 0
      \node[very thick, draw=green] (01) at (0,0) {$h_1^0$};
      \node[draw=blue] (o01) at (0,1.25) {$o_1^0$};

      \node[very thick, draw=green] (02) at (0,4) {$h_2^0$};
      \node[draw=blue] (o02) at (0,5.25) {$o_2^0$};

      \node[very thick, draw=green] (03) at (0,8) {$h_3^0$};
      \node[draw=blue] (o03) at (0,9.25) {$o_3^0$};

      % step 1
      \node[very thick, draw=green] (11) at (5,0) {$h_1^1$};
      \node[draw=blue] (o11) at (5,1.25) {$o_1^1$};

      \node[very thick, draw=green] (12) at (5,4) {$h_2^1$};
      \node[draw=blue] (o12) at (5,5.25) {$o_2^1$};

      \node[very thick, draw=green] (13) at (5,8) {$h_3^1$};
      \node[draw=blue] (o13) at (5,9.25) {$o_3^1$};

      % step 2
      \node[very thick, draw=green] (21) at (10,0) {$h_1^2$};
      \node[draw=blue] (o21) at (10,1.25) {$o_1^2$};

      \node[very thick, draw=green] (22) at (10,4) {$h_2^2$};
      \node[draw=blue] (o22) at (10,5.25) {$o_2^2$};

      \node[very thick, draw=green] (23) at (10,8) {$h_3^2$};
      \node[draw=blue] (o23) at (10,9.25) {$o_3^2$};
  \end{scope}

  \begin{scope}[>={Stealth[black]},
                every node/.style={fill=white, circle, pos=0.80,inner sep=1},
                every edge/.style={draw=black}]

      %STEP 0
      \path [->] (x01) edge (01);
      \path [->] (x02) edge (02);
      \path [->] (x03) edge (03);

      \path [->] (01) edge (o01);
      \path [->] (02) edge (o02);
      \path [->] (03) edge (o03);

      %STEP 1
      \path [->] (x11) edge (11);
      \path [->] (x12) edge (12);
      \path [->] (x13) edge (13);

      \path [->] (11) edge (o11);
      \path [->] (12) edge (o12);
      \path [->] (13) edge (o13);  

      \path [->] (01) edge[bend left=0, dashed] (11);
      \path [->] (01) edge node {$m_{12}^1$} (12);
      \path [->] (01) edge node {$m_{13}^1$} (13);

      \path [->] (02) edge node {$m_{21}^1$} (11);
      \path [->] (02) edge[bend left=0, dashed] (12);
      \path [->] (02) edge node {$m_{23}^1$} (13);

      \path [->] (03) edge node {$m_{31}^1$} (11);
      \path [->] (03) edge node {$m_{32}^1$} (12);
      \path [->] (03) edge[bend left=0, dashed] (13);

      %STEP 2
      \path [->] (x21) edge (21);
      \path [->] (x22) edge (22);
      \path [->] (x23) edge (23);

      \path [->] (21) edge (o21);
      \path [->] (22) edge (o22);
      \path [->] (23) edge (o23);

      \path [->] (11) edge[bend left=0, dashed] (21);
      \path [->] (11) edge node {$m_{12}^2$} (22);
      \path [->] (11) edge node {$m_{13}^2$} (23);

      \path [->] (12) edge node {$m_{21}^2$} (21);
      \path [->] (12) edge[bend left=0, dashed] (22);
      \path [->] (12) edge node {$m_{23}^2$} (23);

      \path [->] (13) edge node {$m_{31}^2$} (21);
      \path [->] (13) edge node {$m_{32}^2$} (22);
      \path [->] (13) edge[bend left=0, dashed] (23);
  \end{scope}
  \end{tikzpicture}
\caption*{Recurrent relational network on a fully connected graph with 3 nodes. Subscripts denote node indices and superscripts denote steps $t$. The dashed lines indicate the recurrent connections. \label{fig:unrolled}}
\end{figure}
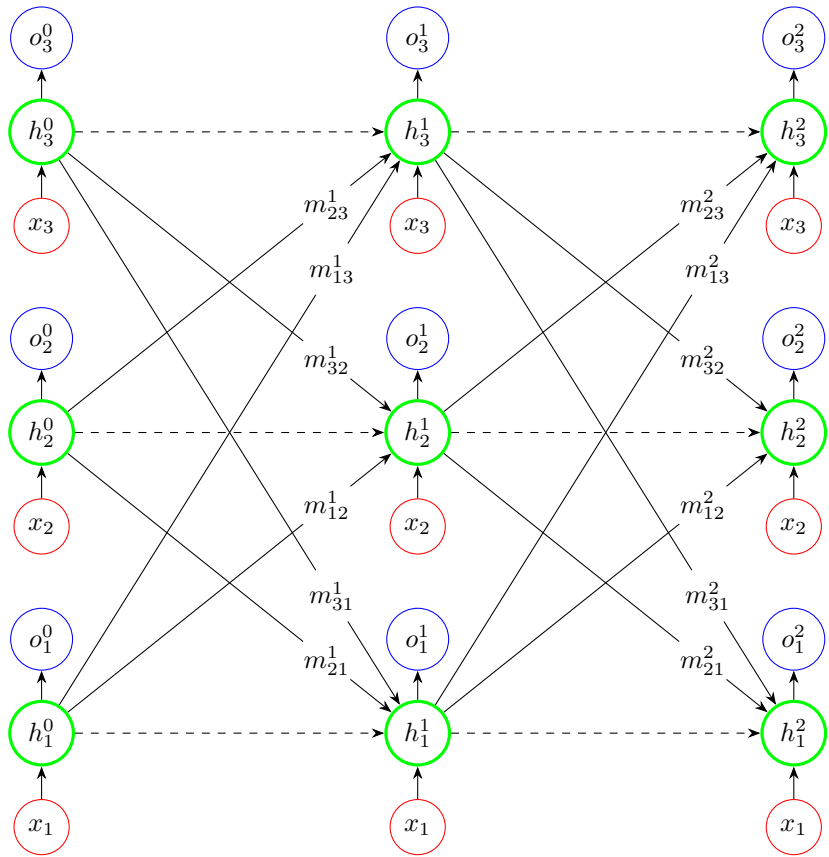

\subsection{Full Sudoku solution}
\begin{figure}
	\caption*{An example Sudoku. Each of the 81 cells contain each digit 1-9, which is useful if the reader wishes to try to solve the Sudoku as they can be crossed out or highlighted, etc. The digit font size corresponds to the probability our model assigns to each digit at step 0, i.e. before any steps are taken. Subsequent pages contains the Sudoku as it evolves with more steps of our model.}
	\centering
	\includegraphics[width=1.0\textwidth]{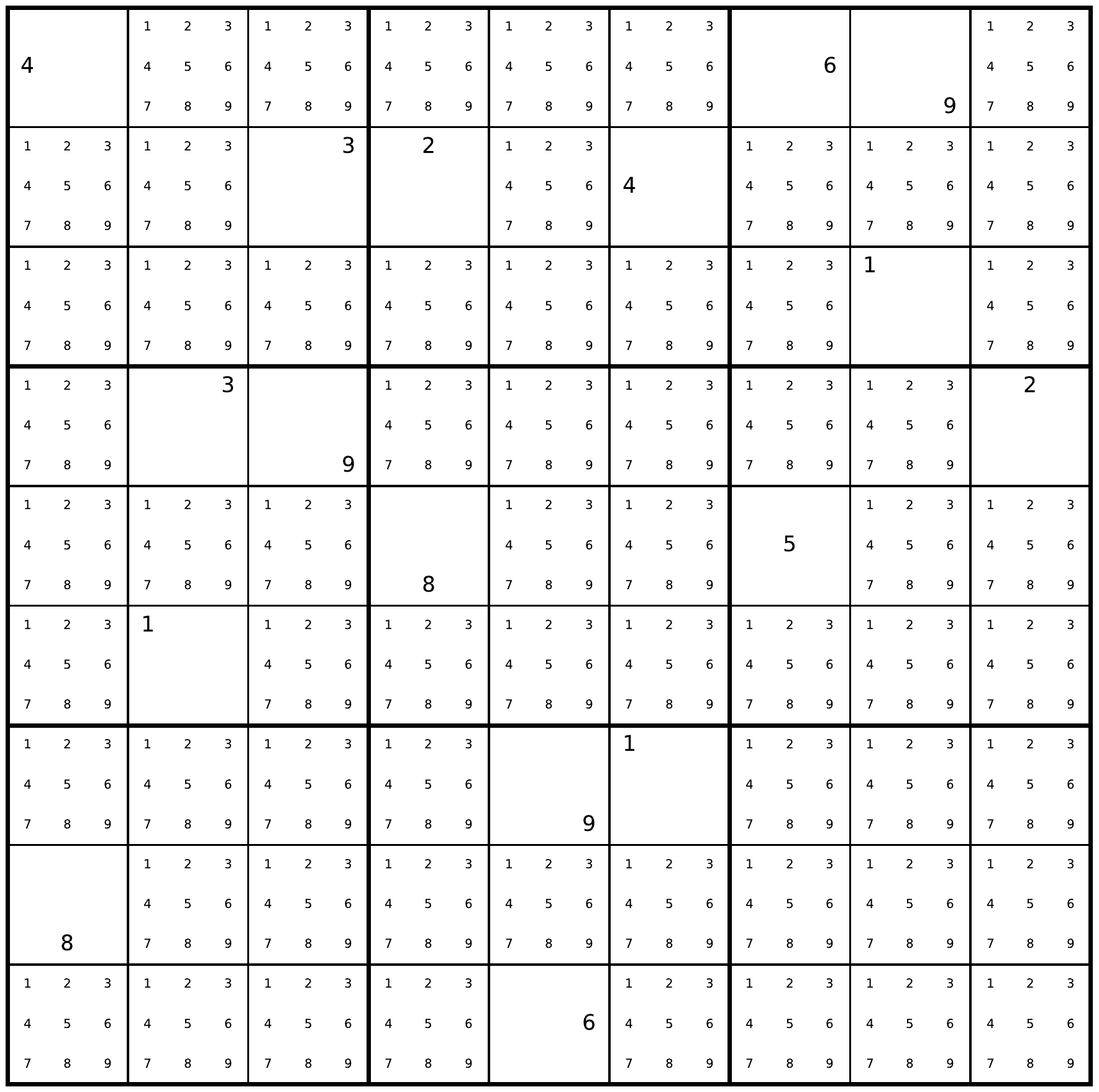}
\end{figure}

\begin{figure}
	\caption*{Step 1}
	\centering
	\includegraphics[width=1.0\textwidth]{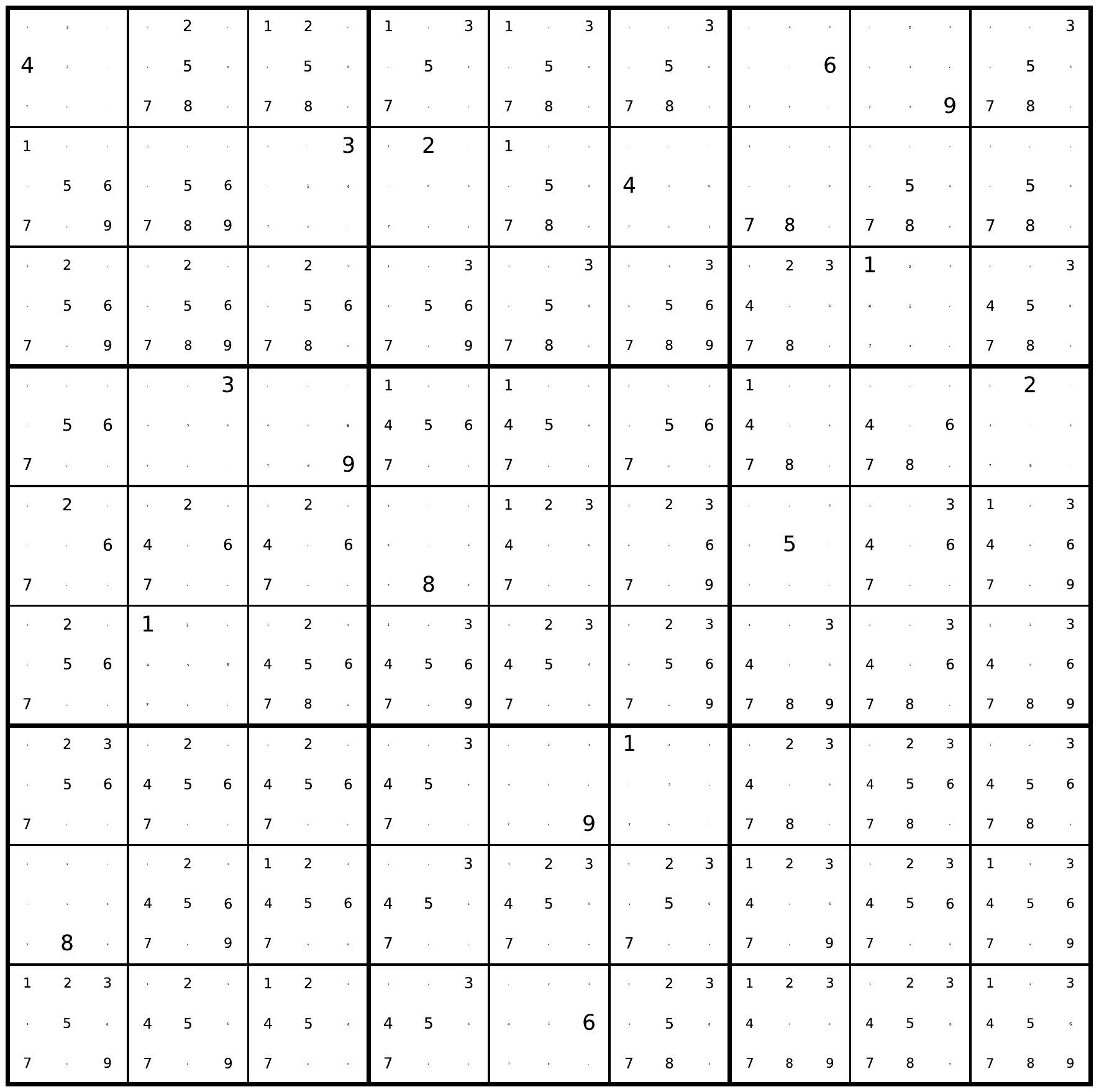}
\end{figure}

\begin{figure}
	\caption*{Step 4}
	\centering
	\includegraphics[width=1.0\textwidth]{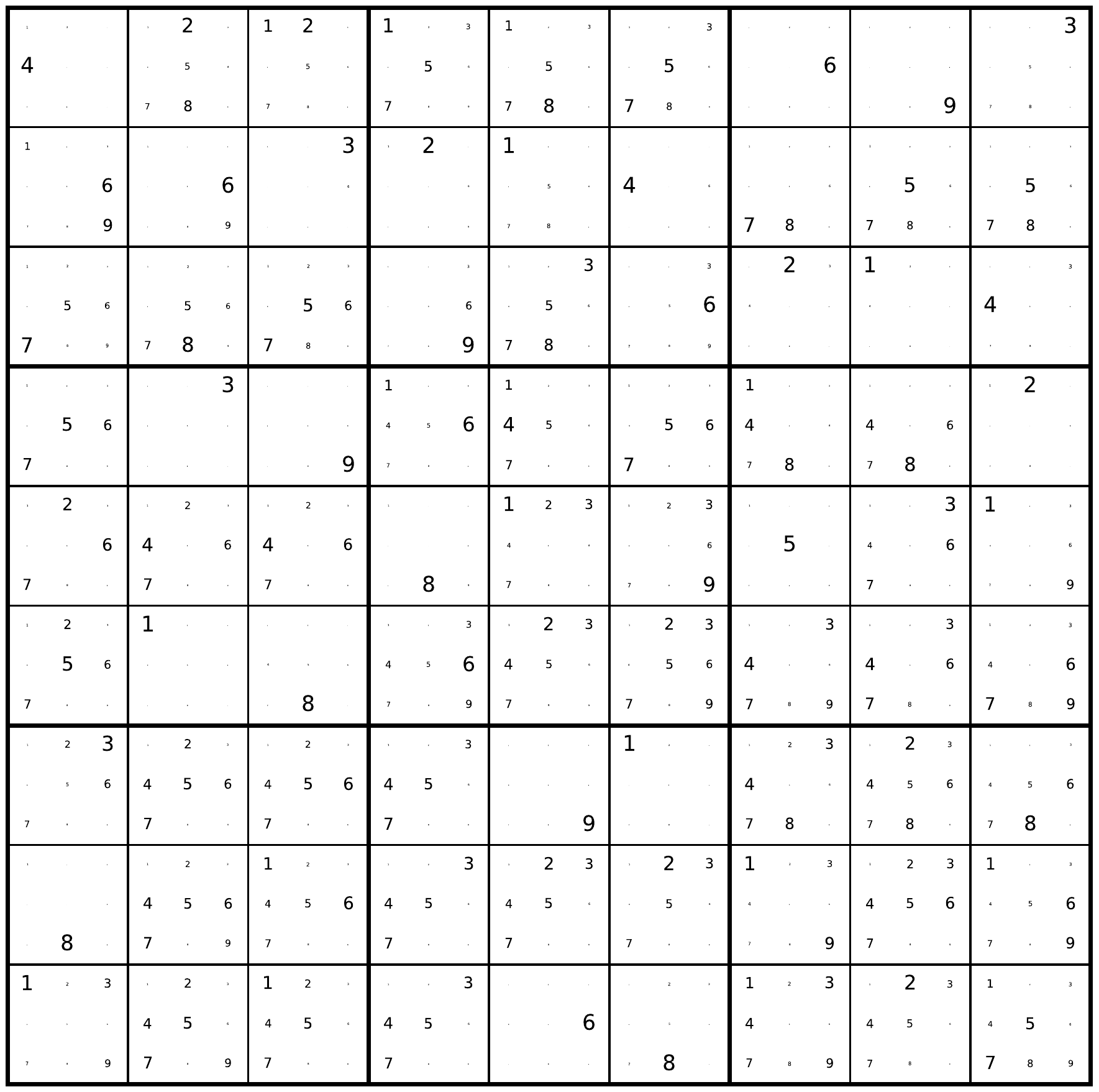}
\end{figure}

\begin{figure}
	\caption*{Step 8}
	\centering
	\includegraphics[width=1.0\textwidth]{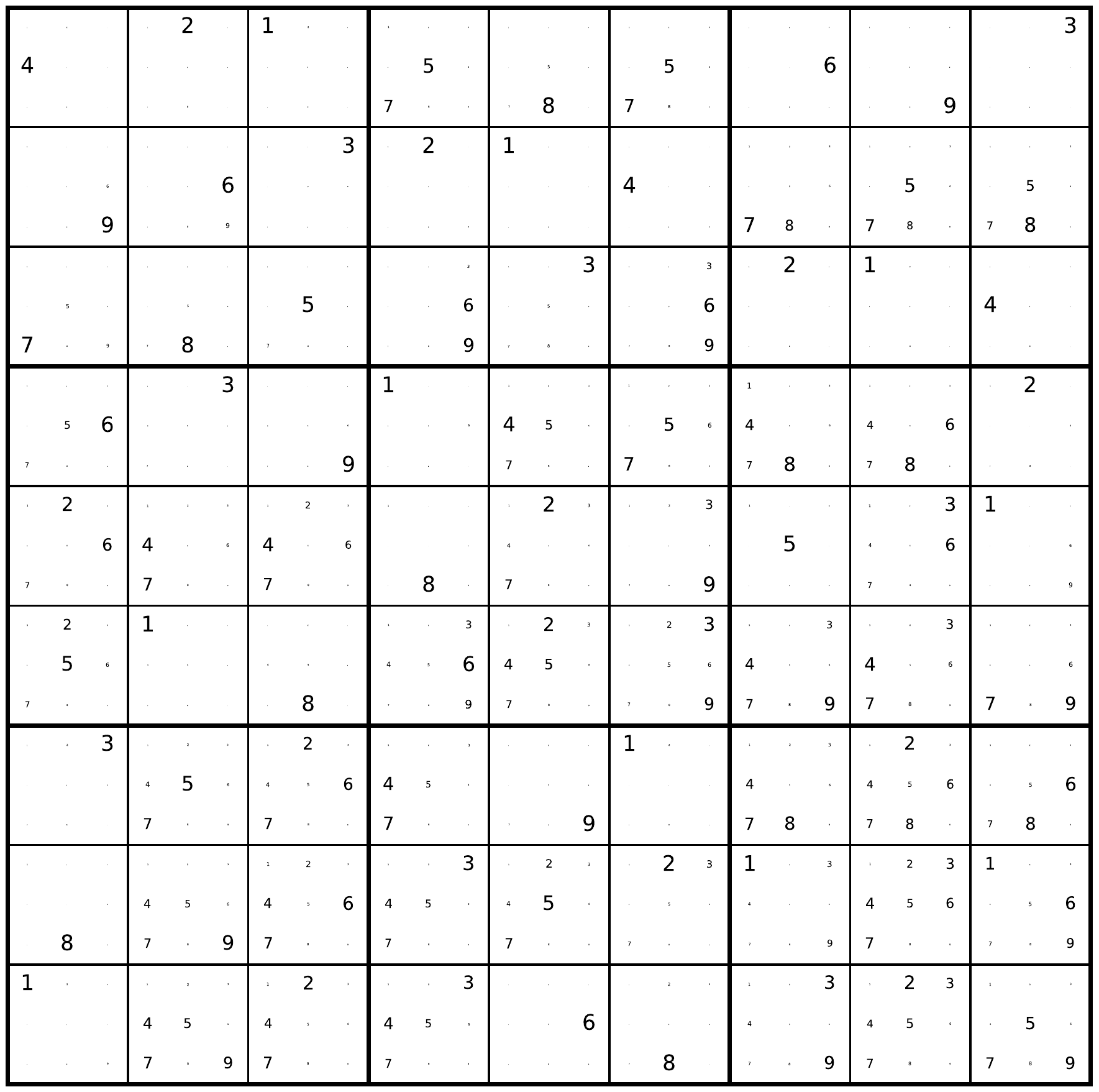}
\end{figure}

\begin{figure}
	\caption*{Step 12}
	\centering
	\includegraphics[width=1.0\textwidth]{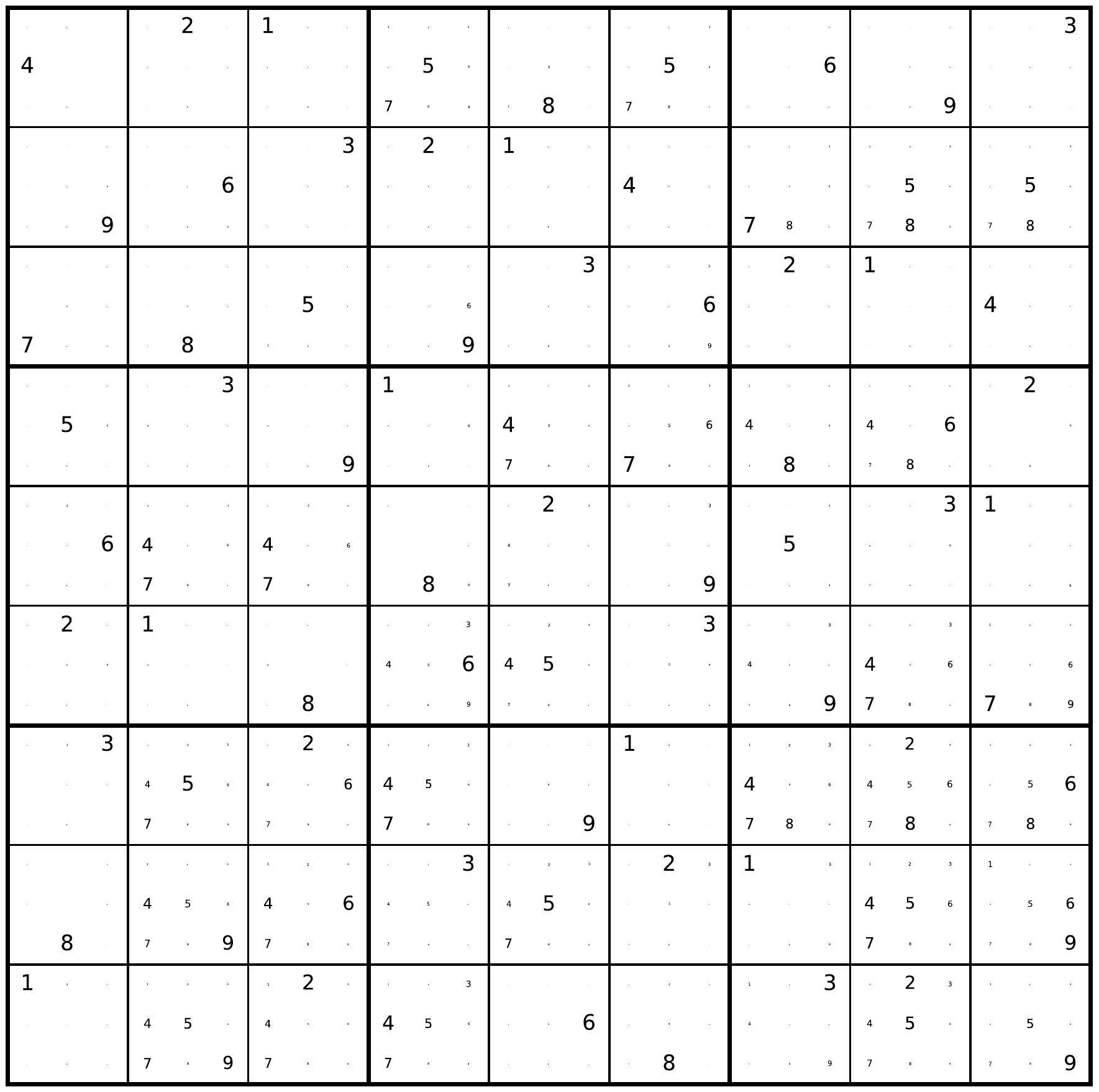}
\end{figure}

\begin{figure}
	\caption*{Step 16}
	\centering
	\includegraphics[width=1.0\textwidth]{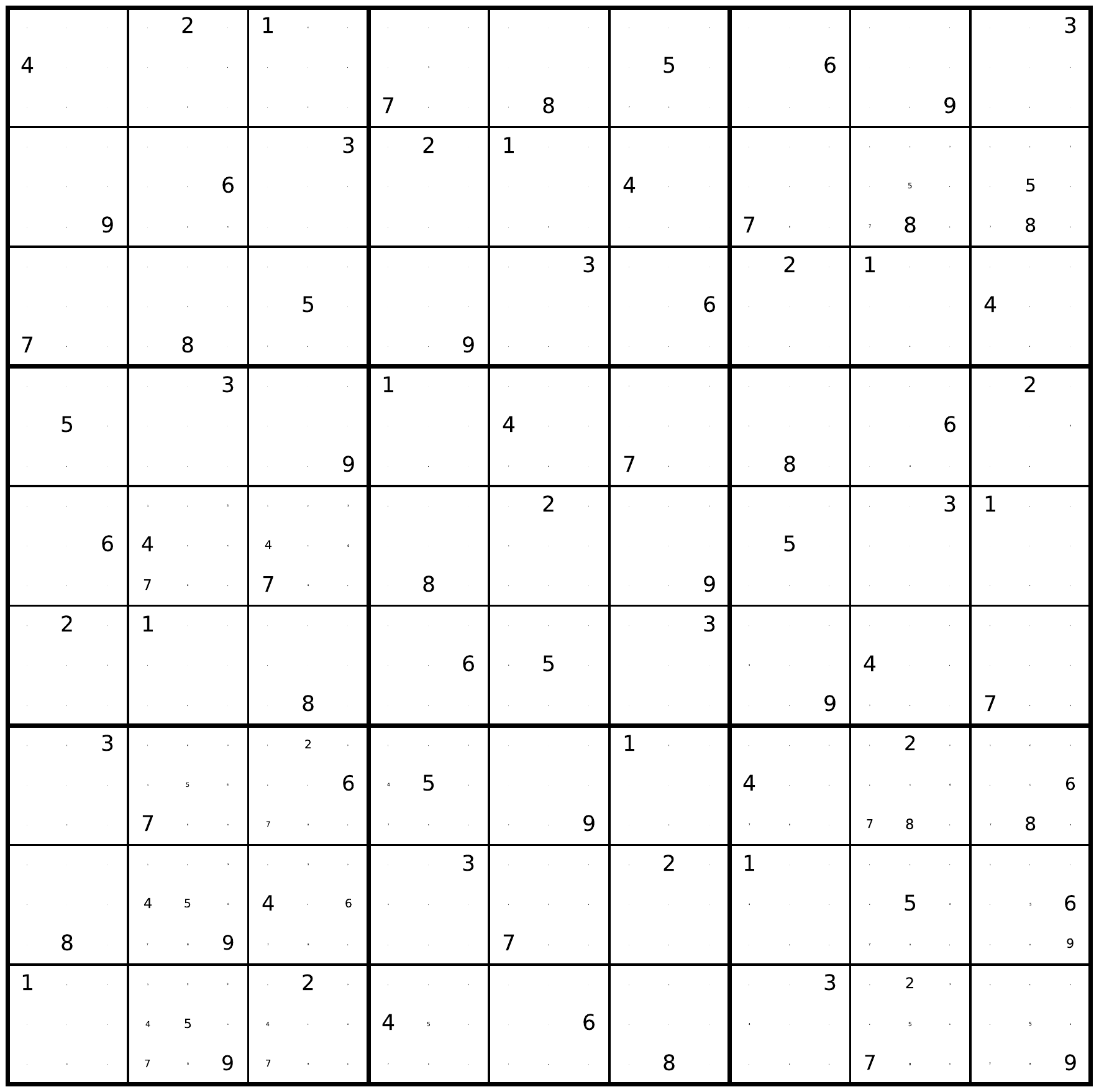}
\end{figure}

\begin{figure}
	\caption*{Step 20}
	\centering
	\includegraphics[width=1.0\textwidth]{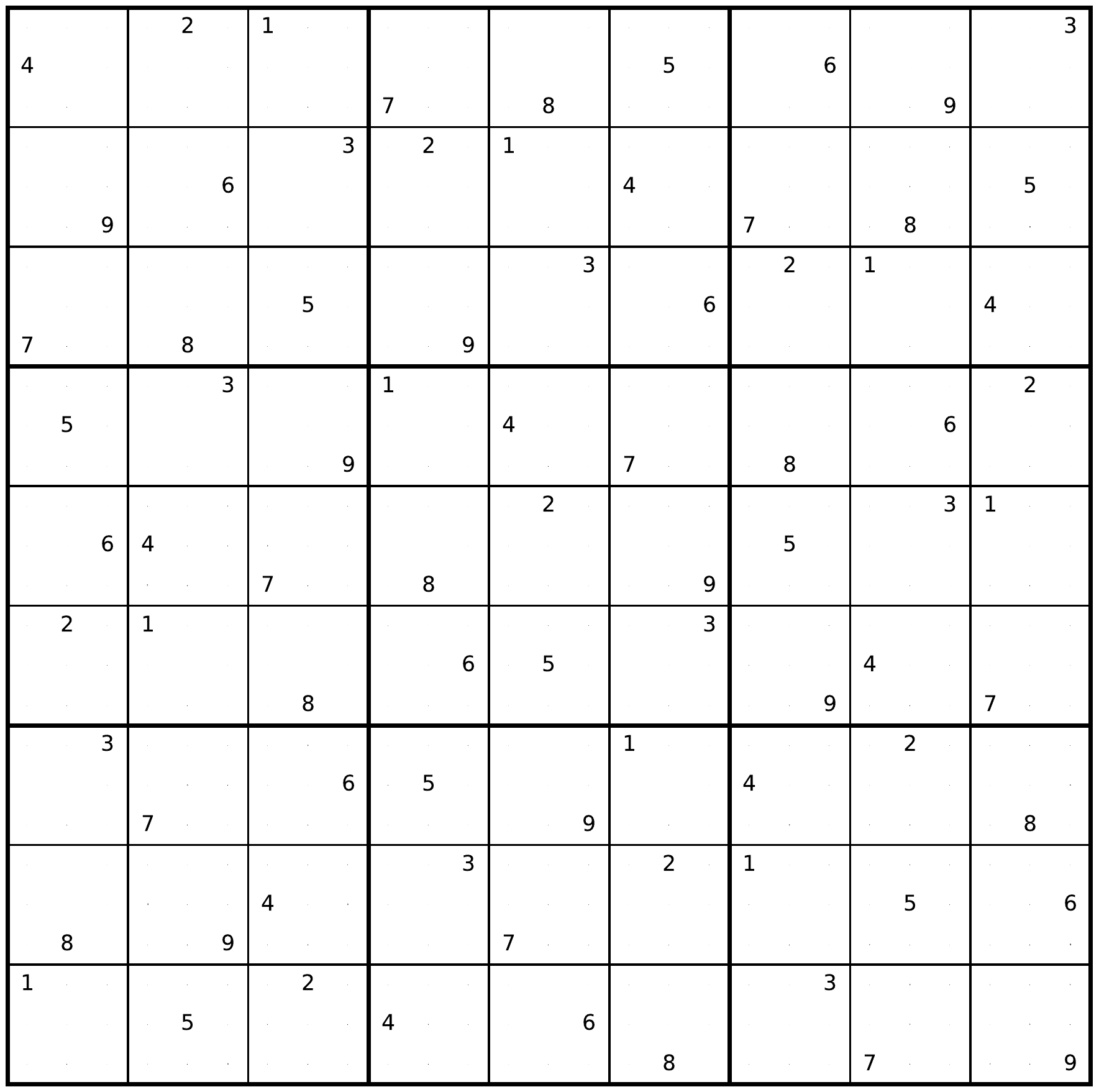}
\end{figure}

\end{document}